# A Simulation Platform for Flapping-Wing Vehicles

Haichuan Li
University of Turku
Email: haichuan.li@utu.fi

Tomi Westerlund
University of Turku
Email: tovewe@utu.fi

*Abstract*—Flapping-wing aerial vehicles (FWAVs) demonstrate remarkable agility but face substantial autonomy challenges due to their high sensitivity to aerodynamic disturbances and limited sensor payload capacity. Current simulation platforms typically rely on oversimplified laminar flow assumptions and idealized sensor models, failing to capture the complex turbulence patterns and perceptual limitations encountered in realworld operation. This simulation-to-reality discrepancy significantly impedes the development of robust autonomy systems for FWAVs. We introduce FWAV-Sim, a high-fidelity Unitybased simulation framework that integrates: (1) a composite aerodynamic model combining quasi-steady blade-element theory with bluff-body drag effects, (2) spatiotemporally correlated turbulence generation through fractal noise synthesis, and (3) realistic sensor simulation including noisy IMU measurements, LiDAR point clouds, and RGB camera feeds. Our platform enables scalable generation of synchronized datasets containing ground-truth vehicle states, aerodynamic forces, turbulent wind fields, and multi-modal sensor streams. Experimental validation demonstrates that autonomy pipelines (including both controllers and perception systems) developed in FWAV-Sim exhibit significantly improved simulation capability, thereby advancing the outstanding performance in simulation-based development for flapping-wing aerial systems.

## I. Introduction

Flapping-wing aerial vehicles (FWAVs) represent a promising paradigm in aerial robotics, offering unprecedented agility in confined spaces and bio-inspired efficiency at low Reynolds numbers [27]. However, their deployment in dynamic environments remains challenging due to three fundamental limitations: (1) extreme sensitivity to unsteady aerodynamic disturbances arising from turbulent airflow [21], and (2) severe payload constraints (3)lack of precise sensor suit that limit onboard sensing capabilities [34]. These challenges are exacerbated by the strong coupling between wing kinematics, body dynamics, perception and environmental disturbances, which demands integrated solutions spanning control, perception, and state estimation.

Existing simulation platforms for aerial robots typically adopt overly simplistic assumptions that fail to capture the complexities of FWAV operation. Multirotor simulators such as AirSim and FlightGoggles focus on photorealistic rendering and idealized sensor models but neglect the unsteady aerodynamics critical to flapping flight. Meanwhile, FWAV-specific simulators [9, 14] often assume laminar flow conditions and provide only limited sensor emulation, creating a significant simulation-to-reality gap when transferring algorithms

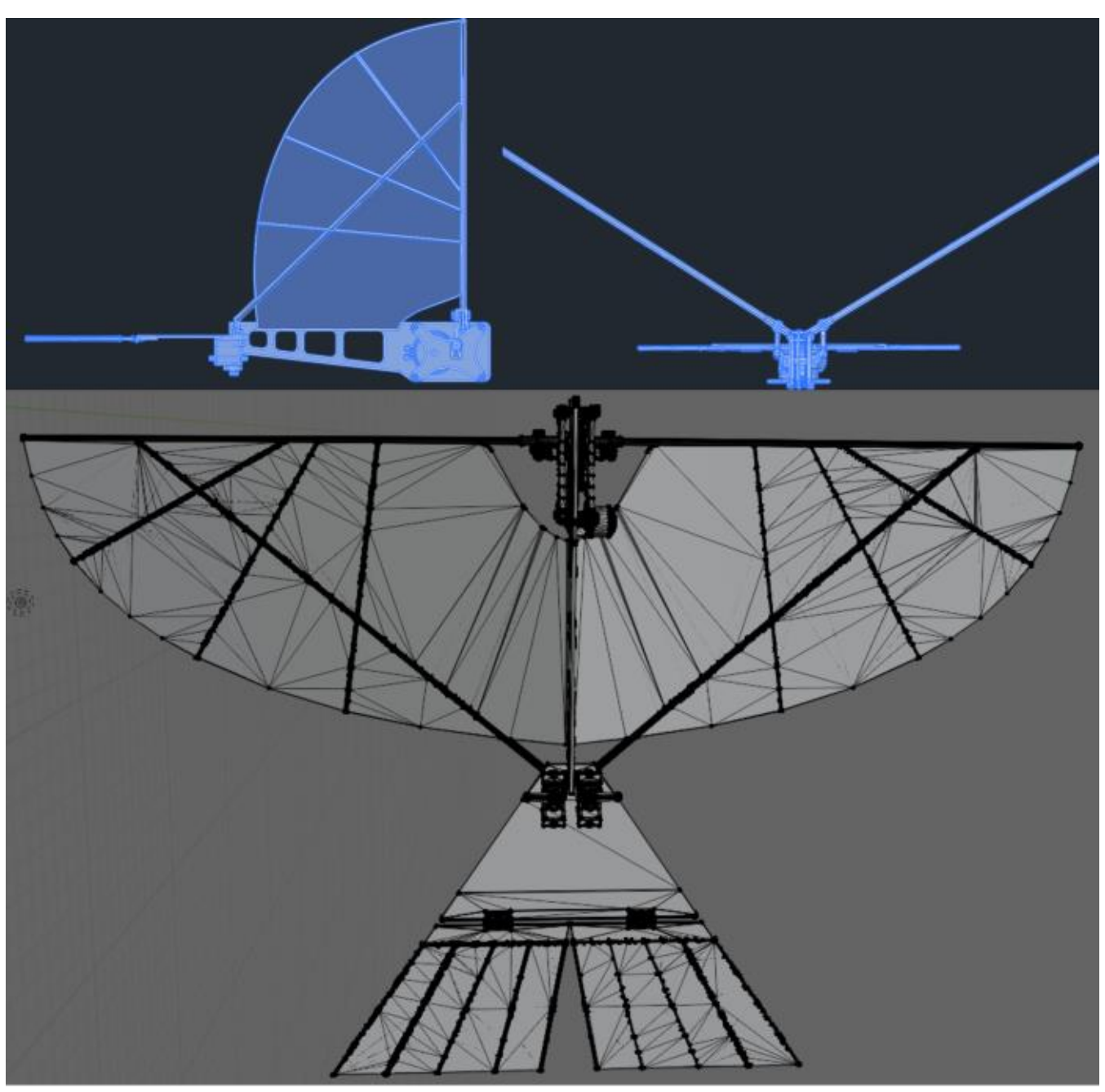

Fig. 1. the mechanical design of one of our bio-inspired FWAVs used in the simulation platform, featuring (top) a 3D model of the complete vehicle structure including the fuselage, wing attachment mechanisms, and sensor mounting points, and (bottom) the detailed mesh of the vehicle for physic interaction

to physical systems. This gap is particularly problematic for data-driven methods, which require large-scale, high-fidelity datasets that capture the interaction between turbulent airflow, vehicle dynamics, and noisy perception.

To address these challenges, we introduce FWAV-Sim, a high-fidelity, real-time simulation framework jointly models:

- A composite aerodynamic system combining quasisteady blade-element theory for unsteady lift/thrust generation with bluff-body drag to capture fuselage-flow interactions,
- A dynamic wind field generator that synthesizes spatiotemporally correlated turbulence via fractal noise, enabling realistic disturbance patterns,
- A multi-modal sensor suite with physically plausible noise models, including IMU, LiDAR, and RGB cameras with motion blur and exposure effects.

Our proposed framework enables scalable generation of synchronized datasets containing ground-truth states, aerodynamic forces, wind field vectors, and noisy sensor streams. This comprehensive data collection capability supports: (i) Development of robust control policies that explicitly account for turbulent disturbances, (ii) Training of perception algorithms under realistic sensor conditions, (iii) Benchmarking of state estimation pipelines that fuse noisy measurements with aerodynamic models, (iv) learning-based method of autonomy stacks that jointly optimize perception and control.

Through extensive validation, we demonstrate that algorithms developed in FWAV-Sim achieve comprehensive ability with control and perception tasks compared to the exist platforms, with particular gains in gust rejection and perception robustness. By providing a unified platform for co-designing control and perception systems under realistic conditions, FWAV-Sim represents a significant step toward high-realistic simulation system for flapping-wing aerial robots.

## II. Related Work

### A. Simulation Platforms for Aerial Robotics

Simulation environments are a cornerstone of aerial robotics research, enabling safe, repeatable, and scalable development of control, perception, and planning algorithms. For multirotor and fixed-wing platforms, a number of mature simulators are widely used, including Gazebo, AirSim [28], and FlightGoggles [10]. These frameworks offer physics-based dynamics, configurable sensor suites, and increasingly photorealistic rendering, and have proven effective for vision-based navigation, learning-based control, and system-level validation. Complementary datasets such as EuRoC MAV [3], Blackbird [1], and UZH-FPV [7] further support algorithm development and benchmarking.

Despite their success, these simulators are not directly applicable to flapping-wing aerial vehicles. Most assume rigidbody dynamics with rotor- or fixed-wing–based force models and do not natively support flapping-wing actuation, unsteady lift generation, or the strong coupling between wing kinematics and body motion [2, 20]. As a result, extending such platforms to FWAVs often requires substantial simplifications that compromise aerodynamic realism and limit their usefulness for controller development and sim-to-real transfer.

Several FWAV-specific simulators have been proposed, but they typically focus on a narrow subset of system behaviors. Some prioritize detailed wing kinematics and aerodynamic modeling while neglecting environmental complexity and sensor simulation [33, 11]. Others embed simplified flappingwing dynamics into general-purpose robotics simulators, often relying on basic disturbance models that do not reflect realistic conditions [14, 9, 24]. Consequently, there remains a need for simulation platforms that balance aerodynamic fidelity, environmental realism, and real-time performance for FWAV research.

### B. Flapping-Wing Aerodynamic Modeling

The aerodynamics of flapping-wing flight have been studied extensively using analytical, numerical, and experimental approaches. Quasi-steady aerodynamic models, inspired by insect flight studies, are commonly used to approximate lift and drag forces generated by flapping wings at a tractable computational cost [8, 27]. These models are often enhanced with empirical corrections to account for effects such as rotational lift, added mass, and wake capture, providing a practical compromise between accuracy and efficiency.

High-fidelity computational fluid dynamics (CFD) methods, including direct numerical simulation and immersed boundary techniques, have been employed to analyze unsteady flow structures around flapping wings [19, 30]. While these approaches yield valuable physical insight, their high computational cost makes them unsuitable for real-time simulation or large-scale controller training. Experimental studies using wind tunnels, force balances, and motion capture systems have further advanced understanding of FWAV aerodynamics [13, 18], but such setups are difficult to reproduce and integrate into closed-loop simulation environments.

As a result, most real-time FWAV simulators adopt quasisteady aerodynamic formulations augmented with simplified body drag models. The challenge lies in capturing the dominant aerodynamic interactions while maintaining sufficient computational efficiency to support real-time simulation and large-scale experimentation. Our approach follows this paradigm by coupling a quasi-steady wing model with a bluffbody drag formulation, enabling physically interpretable force modeling suitable for interactive simulation.

### C. Wind Disturbances and Turbulence Modeling

Modeling wind disturbances remains a critical challenge for aerial vehicle simulation. Many control-oriented studies represent wind as a constant bias, slowly varying disturbance, or parametric gust for robustness analysis [16]. More advanced approaches employ stochastic turbulence models, such as Dryden or von Karm´ an spectra, which are commonly used in´ fixed-wing aircraft analysis [5]. While these models capture certain statistical properties of atmospheric turbulence, they are rarely integrated into real-time robotics simulators with spatially resolved wind fields.

For small aerial vehicles, including MAVs, unsteady and turbulent flows can induce forces comparable in magnitude to those generated by the vehicle itself [21, 31]. Nevertheless, most existing simulation frameworks provide limited support for spatially and temporally varying wind, and often apply wind forces uniformly to the vehicle without distinguishing between interactions with lifting surfaces and the fuselage.

Recent work in learning-based control has explored procedurally generated disturbances, domain randomization,

and synthetic turbulence to improve robustness and simulation-toreality transfer [32, 26, 17]. However, these efforts predominantly target rotorcraft and do not explicitly model flappingwing aerodynamics or the coupled interaction between body motion and airflow [12, 22]. In contrast, our simulation environment generates spatiotemporally varying wind fields using fractal noise and explicitly applies wind-induced forces to both the wings and the vehicle body, enabling more realistic

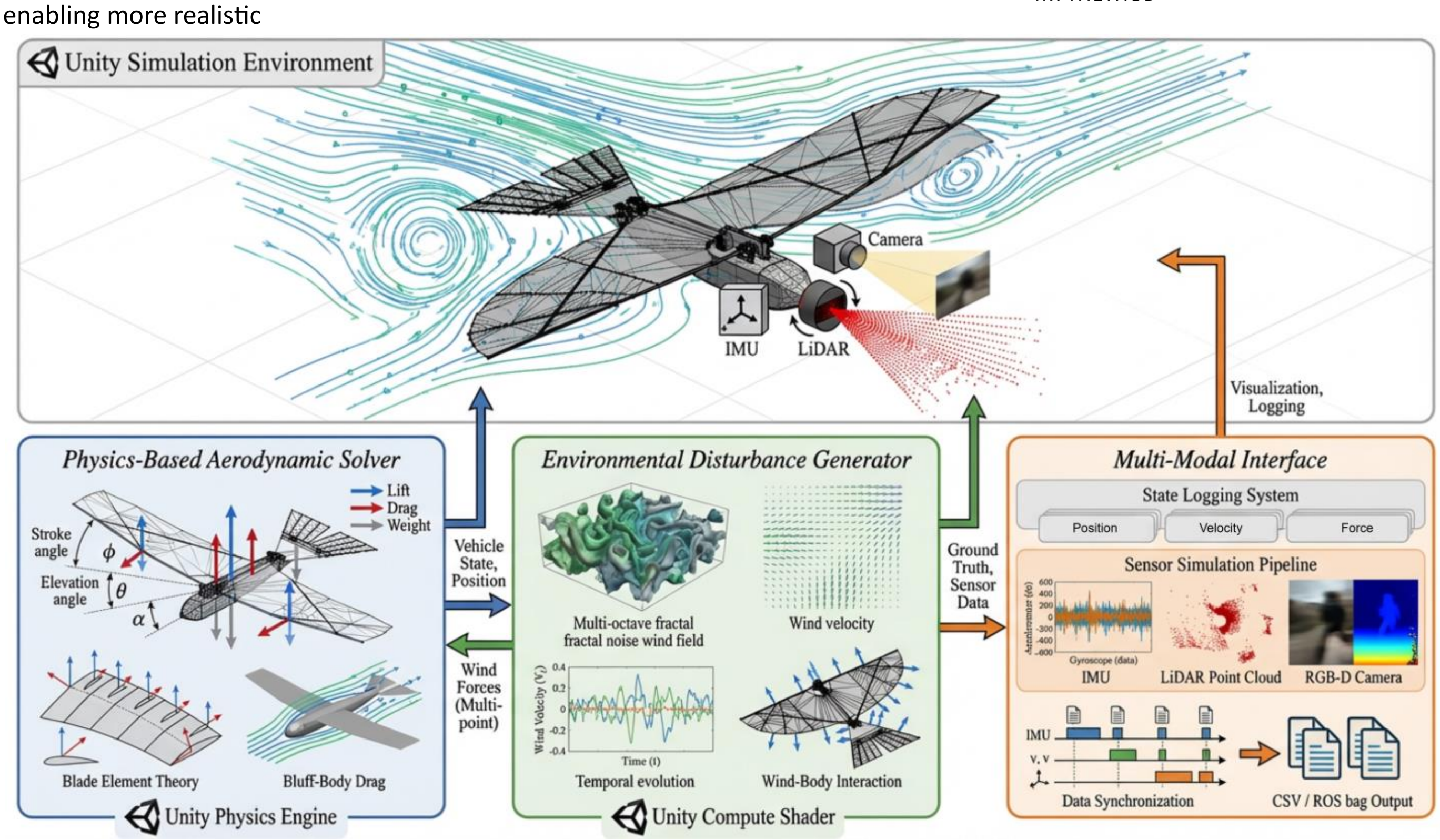

Fig. 2. The integrated system combines (a) physics-based aerodynamic modeling (blade element theory + bluff-body drag) with (b) fractal turbulence generation and (c) multi-modal sensor simulation (IMU, LiDAR, RGB-D) in a Unity environment, enabling high-fidelity analysis of vehicle dynamics and sensor performance in complex wind fields.

and systematic evaluation of FWAV behavior in turbulent conditions.

### D. FWAV Platforms with Sensor Integration

While most flapping-wing aerial vehicle research has focused on aerodynamic modeling, several platforms have incorporated sensor suites to enable advanced autonomy. Physical FWAV prototypes demonstrate varying degrees of sensor integration: the DelFly series [6]incorporates a monocular camera and IMU for optical flow navigation but faces severe payload constraints; Harvard's RoboBee [18] achieves insectscale agility with basic inertial sensors but operates at extreme size limits; and the Griffin platform [24] integrates stereo cameras and IMU for VIO but remains limited to controlled indoor environments. Simulation platforms for FWAVs with sensor capabilities include Flappy [9], which provides basic IMU simulation without realistic noise models or environmental effects; the Griffin Simulator [24], which offers idealized stereo camera and IMU simulation but lacks physics-based sensor noise and disturbance modeling; and MAVLink-based FWAV simulators [14] that feature basic sensor interfaces but with simplified aerodynamics and no environmental coupling.

## III. Method

We present a simulation platform for flapping-wing aerial vehicles (FWAVs) implemented within the Unity physics engine. The platform is designed to model the coupled interaction between rigid-body dynamics, unsteady aerodynamics, and spatiotemporally varying wind disturbances, while maintaining sufficient computational efficiency for interactive simulation and large-scale experimentation. As illustrated in Fig. 2, the system is composed of three tightly integrated modules: (i) a physics-based aerodynamic solver, (ii) an environmental disturbance generator, and (iii) a synchronized multi-modal state and sensor interface.

### A. System Architecture

The simulator advances in discrete physics steps of $\Delta t$ and combines Unity's rigid-body dynamics engine with custom aerodynamic and environmental force models. At each timestep, aerodynamic forces generated by the flapping wings, fuselage drag, and wind-induced loads acting on the vehicle body are computed and applied to the rigid body. The resulting vehicle state is then used to update sensor simulations and environmental configurations.

This modular design enables independent configuration of aerodynamic parameters, wind field characteristics, and sensing modalities, facilitating symbolic evaluation of FWAV behavior under diverse operating conditions.

### B. Physics-Based Aerodynamic Modeling

The FWAV is modeled as a composite rigid body characterized by mass $m$ and inertial tensor $\mathbf{I}$, subject to gravitational forces, inertial reaction moments, and aerodynamic loading. Aerodynamic forces are computed independently for the left and right wings, allowing the simulation to capture asymmetric force generation essential for maneuvering and disturbance rejection.

For each wing surface $k \in L,R$, the instantaneous aerodynamic force $\mathbf{F}_{\text{aero}}^{(k)}$ is computed using a quasi-steady bladeelement approximation:

$$\mathbf{F}_{\text{wing}}^{(k)} = \frac{1}{2}\rho S \left|\mathbf{v}_{\text{eff}}^{(k)}\right|^2 \left[C_L(\alpha_k), \hat{\mathbf{e}}L + C_D(\alpha_k), \hat{\mathbf{e}}D\right], \tag{1}$$

where $\rho$ = 1.225 kg/m$^3$ is the air density under ISA standard conditions and $S$ is the wing planform area. The effective velocity at the wing center of pressure $\mathbf{r}_{cp}$ is given by

$$\mathbf{v}_{\text{eff}}^{(k)} = \mathbf{v}_{\text{body}} + \omega \times \mathbf{r}_{cp} + \mathbf{v}_{\text{wind}}, \tag{2}$$

combining translational velocity, rotational velocity, and the local wind vector sampled from the environmental disturbance field. The unit vectors $\hat{\mathbf{e}}_L$ and $\hat{\mathbf{e}}_D$ denote the lift and drag directions, respectively.

To account for the non-aerodynamic structural drag, a bluffbody drag force is applied to the fuselage center of mass:

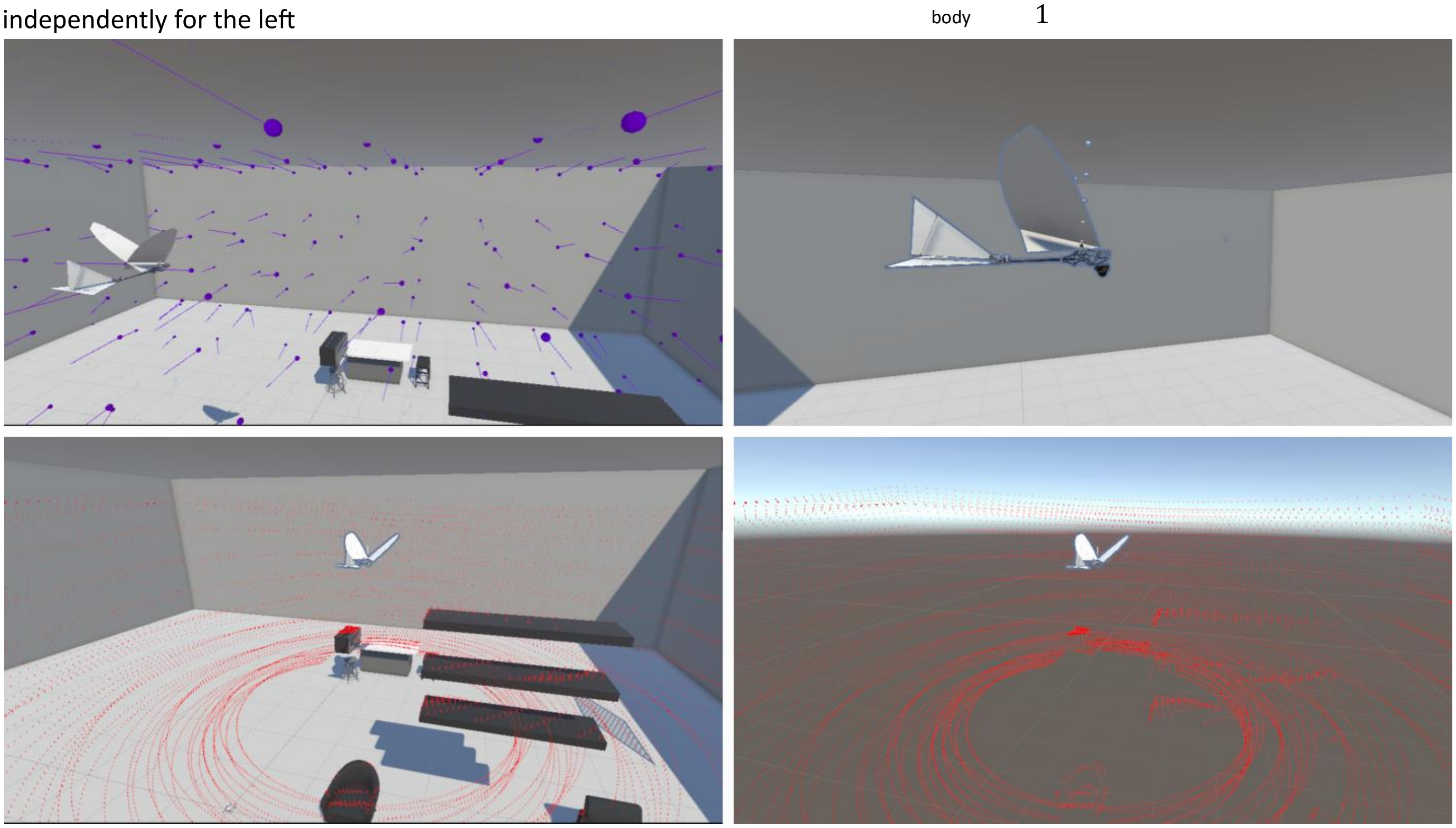

Fig. 3. Multi-modal simulation platform and data collection pipeline for FWAVs. (Top-left: Procedural scene generation) The simulated environment with wind fields and 3D objects, supporting diverse training scenarios. (Top-right: Sensor suite mounted on FWAV) Close-up of the Flapping-Wing Aerial Vehicle (FWAV) equipped with a multi-modal sensor payload, including LiDAR, camera and IMU. (Bottom-left: Active data collection) The FWAV navigates through the procedural environment while logs synchronized wind field interactions, ground-truth states, and noisy sensor measurements. (Bottomright: Collected 3D point cloud) Resulting high-density LiDAR point cloud capturing the vehicle's ego-centric spatial perception of the environment.

$$\mathbf{F}_{\text{drag}}^{\text{body}} = -\frac{1}{2}\rho S_{\text{front}} C_{D,\text{body}} |\mathbf{v}_{\text{rel}}| \mathbf{v}_{\text{rel}} \tag{3}$$

where $S_{\text{front}}$ represents the effective frontal area and $\mathbf{v}_{\text{rel}} = \mathbf{v}_{\text{body}} - \mathbf{v}_{\text{wind}}$ is the relative airspeed.

*a) Nonlinear Angle-of-Attack Dynamics:* Unlike fixedwing aircraft operating at small angles of attack, FWAVs routinely experience large variations in $\alpha$ throughout the flapping cycle. The simulator therefore evaluates the instantaneous angle of attack $\alpha_k$ at every timestep based on the relative orientation between the wing chord and the effective airflow direction.

To capture high-angle aerodynamic behavior, lift and drag coefficients are modeled using nonlinear trigonometric polars:

$$C_L(\alpha) = C_{L\max}\sin(2\alpha),\ C_D(\alpha)$$
$$= C_{D\min} + (C_{D\max} - C_{D\min})(1 - \cos(2\alpha)). \quad (4)$$

This formulation captures stall behavior and force saturation at large angles of attack ($|\alpha| \gg 15^\circ$), enabling realistic force generation during rapid pitch-up and pitch-down phases of flapping flight.

*C. Environmental Disturbance Generation*

To model unstructured airflow, our simulator incorporates a spatially continuous wind field managed by a global wind controller. The local wind velocity experienced by the vehicle is modeled as a superposition of a mean flow component and a turbulent component generated using multi-octave fractal noise(Fig. 4).

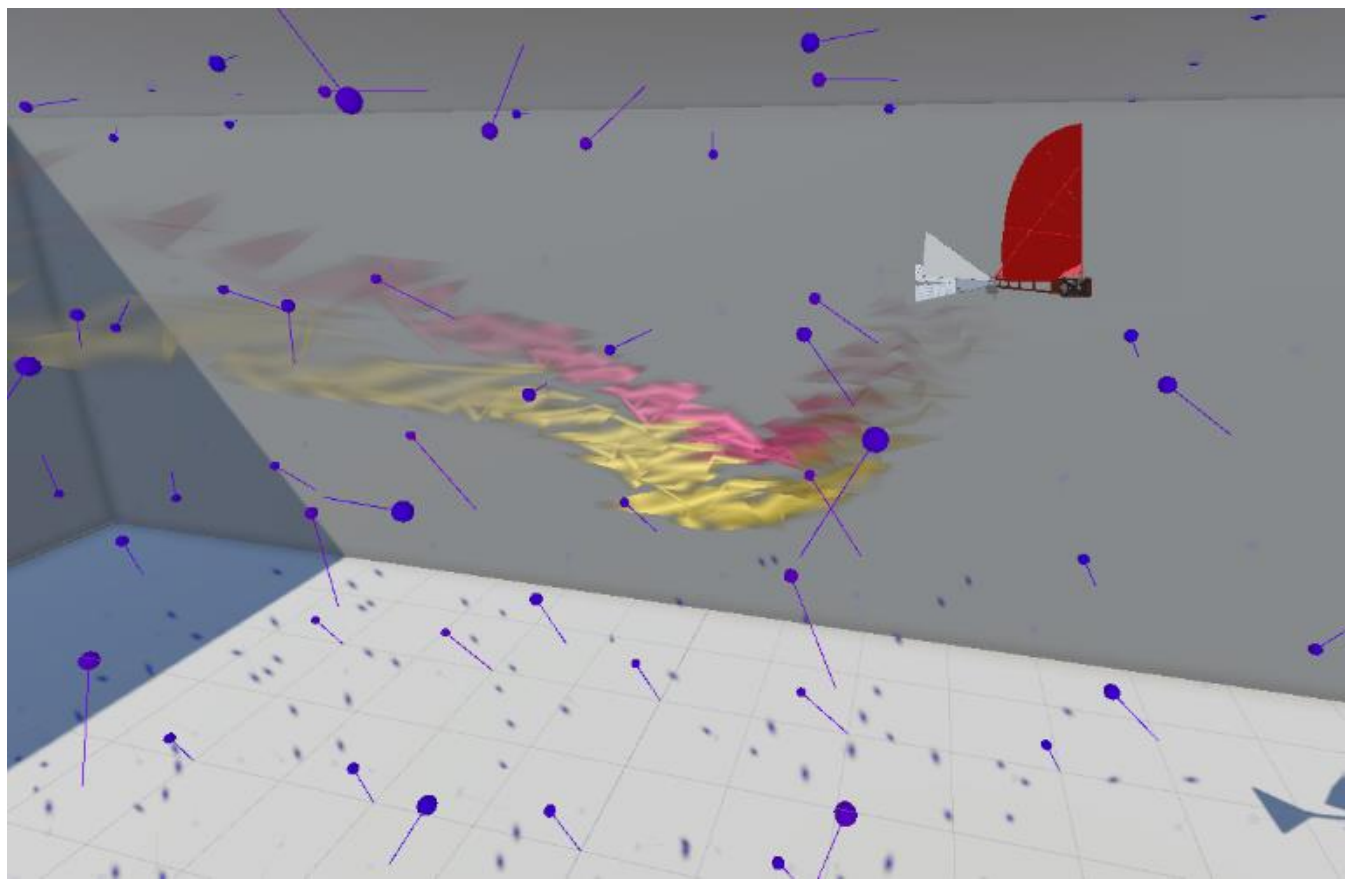

Fig. 4. computational fluid dynamics (CFD) representation of a turbulent airflow field within a confined experimental space, showing the complex wind patterns (purple vectors and vortices) interacting with a flapping-wing micro aerial vehicle (center). The visualization demonstrates the three-dimensional wind velocity vectors and turbulent structures generated for realistic aerodynamic testing, where the vehicle must navigate through spatially-varying, unsteady wind conditions. Such turbulent flow simulations are essential for studying the robust flight control and aerodynamic performance of bio-inspired flapping-wing systems in challenging environmental conditions.

The wind velocity at position $\mathbf{p} = [x,y,z]^T$ and time $t$ is defined as:

$$\mathbf{v}_{\text{wind}}(\mathbf{p},t)\mathbf{v}\text{mean} + \sum^{K} i, \left(\frac{\mathbf{p}}{\lambda s, i} + \mathbf{v}\text{drift}, it\right), \quad i = 1\ \mathbf{A}\ \mathrm{N}! \quad (5)$$

where $\mathbf{v}$mean defines the prevailing wind direction and magnitude, $K$ denotes the number of fractal octaves, and $\mathrm{N}(\cdot)$ is a 3D Perlin noise function producing smooth, $C^1$-continuous turbulence. The parameters $\mathbf{A}i$ and $\lambda s,i$ control the intensity and spatial scale of each turbulence layer, while $\mathbf{v}_{\text{drift},i}$ governs the temporal evolution of gust structures.

This approach yields spatially correlated, time-varying wind disturbances that more closely resemble realistic turbulence than uniform or white-noise models, while remaining deterministic and repeatable across simulation runs.

*D. Sensor Suit*

The perception subsystem in FWAV-Sim addresses the unique sensory challenges inherent to flapping-wing aerial vehicles, which operate under severe payload constraints while experiencing high-frequency vibrations from wing motion. Unlike conventional aerial platforms, FWAV perception systems contend with the aerodynamic coupling between vehicle motion and sensor measurements. Unlike in the real-world devices which are usually light-weight and compact space design, thus carrying heavy or multi sensors is a challenge. Our simulated sensor suit can keep almost same function but light or zero weighted.

The inertial measurement unit (IMU) simulation implements a stochastic noise model capturing standard MEMS sensor characteristics. Accelerometer and gyroscope measurements incorporate Gaussian white noise terms:

$$\mathbf{a}\text{meas} = \mathbf{a}\text{true} + \mathbf{n}a,\ \omega\text{meas} = \omega\text{true} + \mathbf{n}\omega \quad (6)$$ where

$\mathbf{n}a \sim \mathrm{N}(0,\sigma_a^2)$ with $\sigma_a = 0.1$ m/s² and $\mathbf{n}\omega \sim \mathrm{N}(0,\sigma_\omega^2)$ with $\sigma_\omega = 0.05$ rad/s represent the noise densities.

RGB camera model accounts for the dynamic operating conditions of FWAVs. The camera simulation incorporates motion blur through exposure-time-dependent convolution kernels, with the blur direction and magnitude determined by the instantaneous angular velocity vector. Rolling shutter effects are modeled by applying temporal offsets to different image rows, with the offsets synchronized to the wing beat phase to accurately represent the interaction between flapping motion and image acquisition. The simulation further implements dynamic exposure control to handle the varying lighting conditions encountered during aggressive maneuvers, with exposure times automatically adjusted based on scene brightness while respecting the defined frame rate limitations of micro-camera modules. All visual sensors operate with realistic field-of-view constraints (60-90° horizontal) and experience positional noise of $\mathrm{N}(0,0.1\text{mm}^2)$ to represent mechanical vibrations from the flapping mechanism.

LiDAR model generates dense point clouds through parallelized raycast operations. Our LiDAR sensor provides instantaneous $360^\circ$ scans with a programmable update rate. Measurement uncertainty is modeled as additive Gaussian noise $\mathrm{N}(0,\sigma^2)$ applied to the range measurements. Unlike standard hardware, our simulator leverages the simulation state to provide pixel-perfect ground truth semantics. As the point cloud is generated, the pipeline automatically associates each point with its corresponding object instance, extracting 3D oriented bounding boxes and object class labels. This auto-labeling capability eliminates the need for manual data annotation, significantly accelerating the training of 3D object detection and semantic segmentation networks.

A defining characteristic of the FWAV-Sim perception system is its modeling of the complex interactions between aerodynamic disturbances and sensor measurements. Our simulator propagates wind-induced vibrations through the vehicle's mechanical structure to all sensors, creating time-varying noise profiles in inertial measurements that correlate with turbulent airflow conditions. Optical flow patterns in camera images accurately reflect the true airspeed relative to the wind field, with the flow vectors modulated by both vehicle motion and wind velocity. The LiDAR simulation accounts for atmospheric conditions and vehicle attitude in determining point cloud density, with reduced return rates in turbulent conditions and when operating at high angles of attack. All sensory modalities operate within a fully synchronized pipeline that maintains precise temporal alignment with the physics engine's update rate, enabling accurate sensor fusion and state estimation algorithm development.

### *E. Simulation Protocol and Data Logging Architecture*

Our simulation implements an episodic trial system with configurable parameters for environmental conditions, wind disturbance profiles, and vehicle specifications. Each simulation episode follows a standardized protocol comprising initialization, execution, and data logging phases to ensure reproducibility and facilitate comparative analysis. During execution, the system maintains strict temporal synchronization between the physics engine, aerodynamic solver, environmental disturbance generator, and sensor simulation modules, enabling precise correlation between all simulated quantities.

The data logging architecture employs a dual-format storage system optimized for both human readability and machine processing. Physical states and aerodynamic quantities including six-degree-of-freedom pose $(\mathbf{p},\mathbf{q})$, linear and angular velocities $(\mathbf{v},\omega)$, wing kinematics $(\phi,\theta,\psi)$, aerodynamic forces $(\mathbf{L},\mathbf{D})$, and local wind vectors $\mathbf{v}_{wind}$ are serialized in comma-separated value format with microsecond-precision timestamps. Perception data streams comprising LiDAR point clouds and RGB images are stored in efficient binary formats to accommodate their higher data volumes while maintaining temporal alignment with state data.

Our platform features native integration with ROS2 through a dedicated bridge interface that publishes all sensor topics during simulation alongside ground truth topics. This interface supports both simulated and hardware-in-the-loop operation modes, enabling direct integration with standard ROS2 navigation stacks and perception pipelines.

## IV. Experiments

We evaluate the proposed FWAV-Sim platform through a series of experiments designed to assess its realism, versatility, and suitability for developing and benchmarking aerial robotics algorithms. Rather than introducing new algorithmic contributions, the experiments focus on demonstrating that the simulator (i) produces physically meaningful aerodynamic behavior under turbulent wind conditions, (ii) supports perception and state-estimation pipelines commonly used in robotics, and (iii) enables systematic comparison of control strategies under repeatable yet stochastic disturbances.

The evaluation is organized into three categories: perception, control, and odometry.

### *A. Perception Experiments*

The perception experiments assess whether FWAV-Sim provides visually and geometrically realistic sensor data suitable for training and evaluating modern perception algorithms under challenging flight conditions. All experiments are conducted using the simulator's RGB camera and LiDAR sensors, synchronized with the physics and wind simulation.

*1) Semantic Segmentation:* We evaluate four representative semantic segmentation architectures on RGB images rendered by FWAV-Sim: (1) U-Net [25], (2) DeepLabV3+ [4] with an Xception backbone, (3) PSPNet [39] with a ResNet-50 backbone, and (4) SegFormer [35] (MiT-B2 variant). These models span convolutional and transformer-based designs with varying computational complexity.

As shown in Table I, all models achieve reasonable segmentation performance, indicating that the rendered imagery exhibits sufficient visual realism and semantic consistency. SegFormer achieves the highest accuracy while maintaining the lowest computational cost, suggesting that FWAV-Sim can support evaluation of real-time perception pipelines suitable for resource-constrained platforms.

TABLE I
Semantic segmentation performance on FWAV-Sim RGB data

| Model | mAP | mIoU | Pixel Acc. | Params (M) | GFLOPs |
|---|---|---|---|---|---|
| U-Net | 0.68 | 0.76 | 0.90 | 31.0 | 55.7 |
| DeepLabV3+ | 0.75 | 0.81 | 0.93 | 41.2 | 82.5 |
| PSPNet | 0.74 | 0.80 | 0.92 | 49.1 | 96.3 |
| SegFormer | 0.77 | 0.82 | 0.94 | 24.7 | 20.1 |

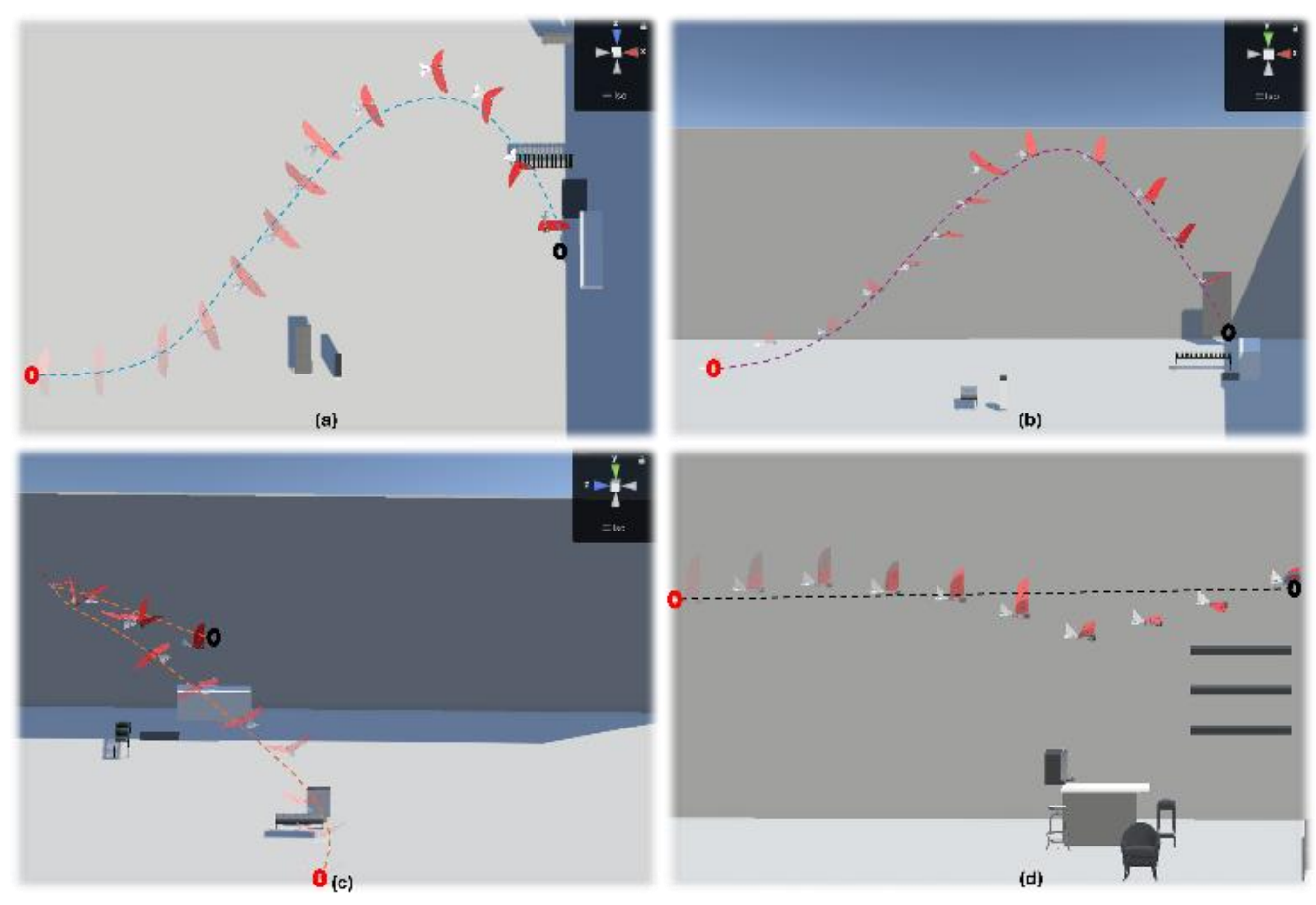

Fig. 5. Trajectory tracking of a RL-based controller.

*2) 3D Object Detection:* To evaluate the geometric fidelity of LiDAR data generated by FWAV-Sim, we benchmark four established 3D object detection architectures: PointPillars [15], SECOND [37], PV-RCNN [29], and CenterPoint [38]. Models are trained to detect furniture, dynamic objects, and static obstacles.

TABLE II
3D OBJECT DETECTION PERFORMANCE (AP@0.5 IOU)

| Model | Furniture | Dynamic | Obstacles | mAP |
|---|---|---|---|---|
| PointPillars | 0.71 | 0.63 | 0.74 | 0.69 |
| SECOND | 0.74 | 0.66 | 0.76 | 0.72 |
| PV-RCNN | 0.77 | 0.69 | 0.79 | 0.75 |
| CenterPoint | 0.79 | 0.72 | 0.81 | 0.77 |

The results demonstrate that the simulated LiDAR data supports state-of-the-art 3D perception pipelines, with performance trends consistent with those reported on real-world datasets. This indicates that FWAV-Sim provides geometrically coherent point clouds suitable for algorithm development and evaluation.

*B. Control Experiments Under Turbulent Wind*

We next evaluate FWAV-Sim as a testbed for control algorithm benchmarking under controlled but realistic wind disturbances. The objective is not to propose a new controller, but to demonstrate that the simulator enables meaningful differentiation between control strategies as wind intensity increases.

*1) Experimental Setup:* Four representative control approaches are evaluated: (1) PID control with gain scheduling, (2) nonlinear model predictive control (NMPC), (3) a reinforcement learning policy trained with PPO, and (4) L1 adaptive control. All controllers are tasked with tracking a figure-eight trajectory while exposed to mean wind speeds ranging from 0 to 5 m/s with superimposed turbulent gusts. Performance is evaluated using:

- Position Tracking Error ($E_{\text{pos}}$): RMSE between the ground-truth position and the reference trajectory.
- Control Effort ($J_{\text{kin}}$): Integrated squared wing kinematics, measuring actuation demand.
- Robustness (R): Success rate without violating attitude or altitude constraints.

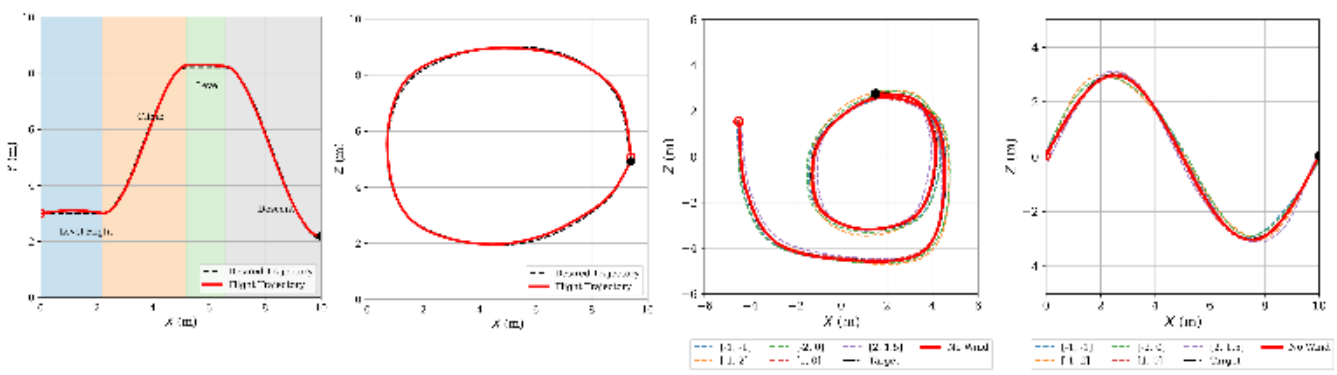

Fig. 6. Reference trajectory and example FWAV trajectories under different control commands.

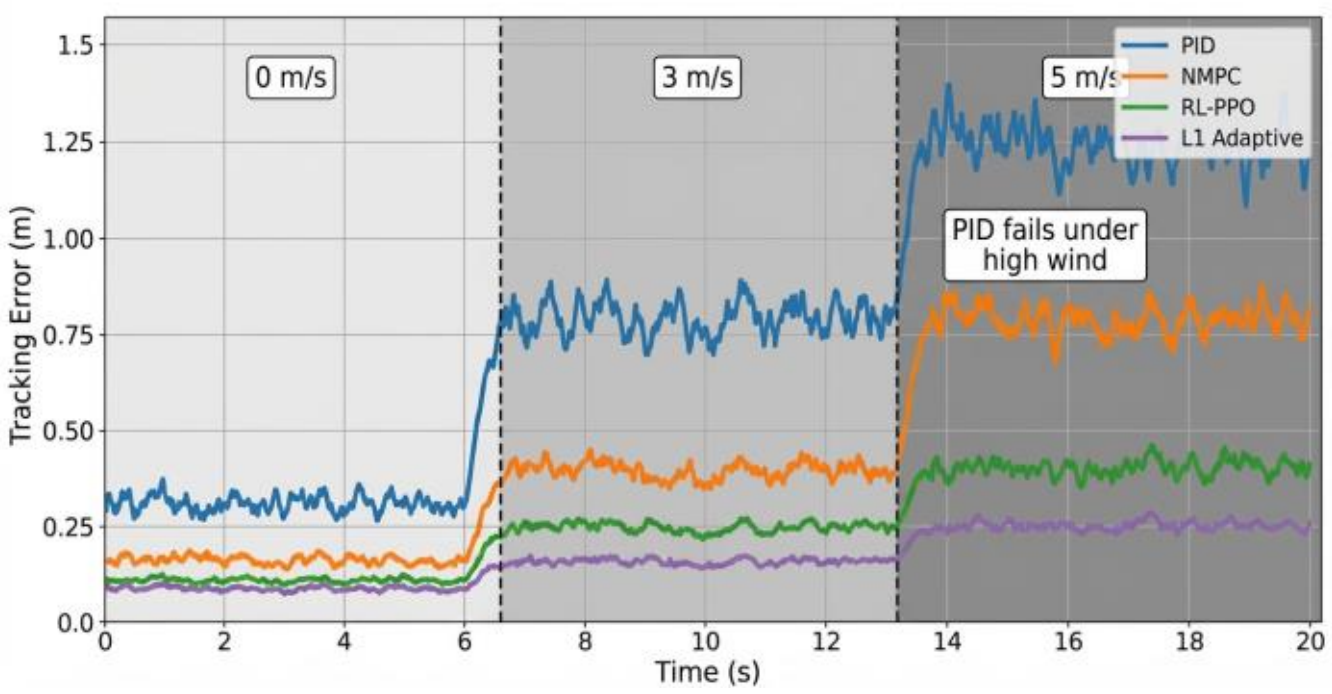

Fig. 7. Tracking error over time under different wind conditions.

TABLE III
CONTROL PERFORMANCE UNDER INCREASING WIND INTENSITY

| Controller | Wind | RMSE (m) | Effort | Success |
|---|---|---|---|---|
| PID | 0 | 0.42 | 1.8 | 0.95 |
| | 3 | 0.68 | 2.5 | 0.72 |
| | 5 | 1.21 | 3.8 | 0.41 |
| NMPC | 0 | 0.18 | 1.2 | 1.00 |
| | 3 | 0.35 | 1.9 | 0.89 |
| | 5 | 0.72 | 2.8 | 0.63 |
| RL (PPO) | 0 | 0.15 | 1.1 | 1.00 |
| | 3 | 0.22 | 1.4 | 0.98 |
| | 5 | 0.31 | 1.7 | 0.92 |
| L1 Adaptive | 0 | 0.14 | 1.1 | 1.00 |
| | 3 | 0.19 | 1.3 | 0.99 |
| | 5 | 0.24 | 1.5 | 0.98 |

*2) Results:* The results show that FWAV-Sim consistently exposes performance differences between controllers as wind disturbances increase. Adaptive and learning-based controllers maintain stable performance under high turbulence, while model-based and classical approaches degrade more rapidly.

This confirms that the simulator provides a sufficiently challenging and physically meaningful environment for controller evaluation.

### C. Force-Related Experimental Methodology

Physics-Informed State Space Formulation. Our experimental methodology formulates the force prediction problem using a physics-grounded state space representation derived from the simulation environment. The system outcomes are encoded as 4-dimensional force vectors:

$\mathbf{F}_{target} = [L_L, D_L, L_R, D_R]^T$ where $L$ and $D$ represent the instantaneous lift and drag forces computed by the blade-element solver (sampled at 50Hz). The signed vertical lift formulation employs dot product decomposition relative to the body frame:

$$L_{signed} = \mathbf{L}_{aero} \cdot \hat{\mathbf{k}}_{body}$$

to distinguish upstroke phases ($L_{signed} < 0$) from downstroke phases ($L_{signed} > 0$).

Virtual Data Collection Protocol. Data is collected using a programmatic flight scenario orchestrator which synchronizes:

- Kinematics: Exact Euler angle extraction from the rigidbody hierarchy (Simulated Motion Capture).
- Dynamics: Aerodynamic forces logged directly from the physics engine step.
- Environment: Ground-truth local wind vectors extracted from the global WindManager.

To ensure adequate domain coverage, our simulation systematically varies: (1)Flapping Frequency: 10-30Hz (Randomized per episode), and (2)Wind Conditions: 0-5m/s (via 3D Perlin Noise) resulting in a dense dataset covering the operational flight envelope.

Biomechanical Action Space Parameterization. The wing kinematic representation uses the standard Euler triplet $\mathbf{u} = [\phi, \theta, \psi]^T$ logged from the Unity Transform hierarchy. The signals are wrapped to a continuous domain $[-\pi, \pi]$ to facilitate learning, corresponding to the physical axes:

- Stroke ($\phi$): Rotation about local X-axis.
- Elevation ($\theta$): Deviation about local Y-axis.
- Pitch ($\psi$): Incidence change about local Z-axis.

Unlike physical markers which suffer from occlusion, our simulation guarantees stable kinematic labels with floatingpoint precision.

Physics-Informed Neural Network Training. We employ a hybrid neural network architecture that combines: 3 fullyconnected layers (64 units each) with Swish activation , Physics-informed output layer that projects to constrained action space and Custom loss function combining: MSE for force reproduction accuracy, Constraint violation penalty (weight: 0.1), and Temporal smoothness regularization (weight: 0.01). The model is trained using the Adam optimizer with learning rate $10^{-4}$ and batch size 256 on our collect data.

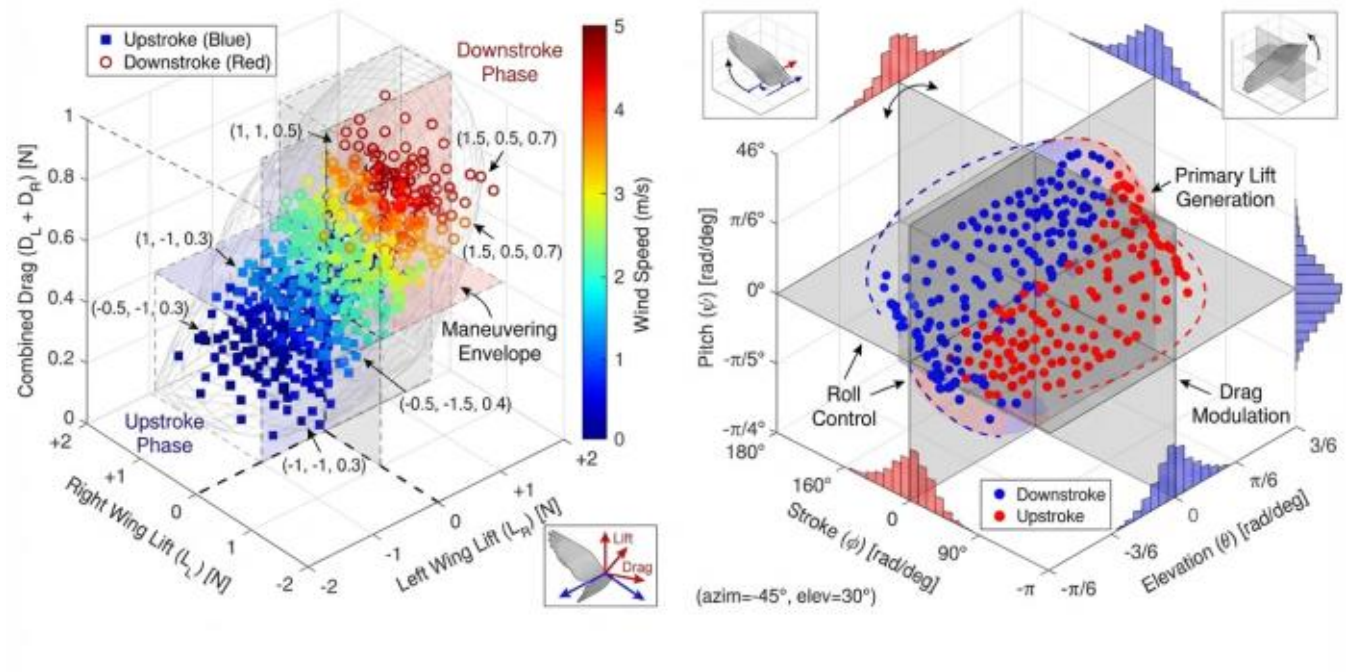

Fig. 8. (Left) Physics-based visualization of the 4-dimensional force vector space in our dataset, showing the signed lift representation that captures the aerodynamic forces during both upstroke and downstroke phases of flapping flight. (Right) The biomechanically-constrained kinematic action space representation for wing configurations, showing the three Euler angles that parameterize the physically-plausible wing states.

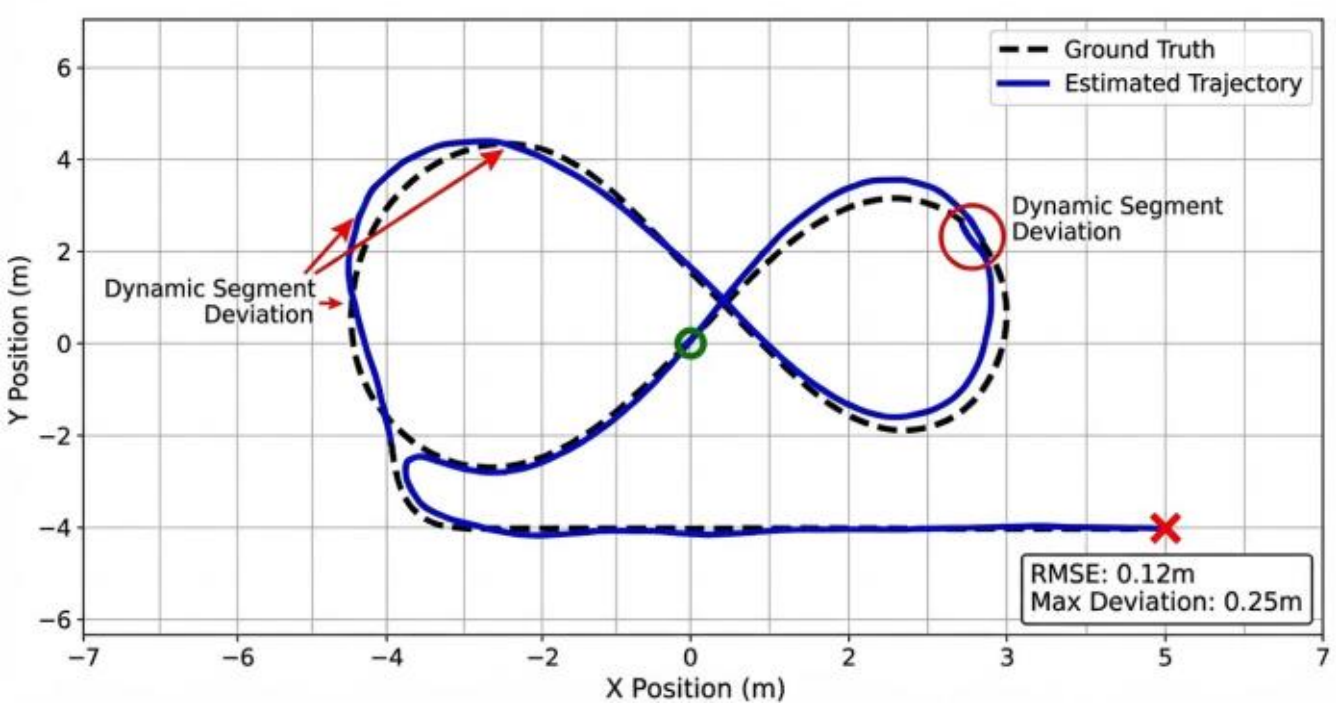

Fig. 9. Ground truth and estimated trajectories for LiDAR–inertial odometry.

### D. Odometry Experiments

Finally, we evaluate whether FWAV-Sim supports realistic state-estimation experiments by benchmarking visual–inertial and LiDAR–inertial odometry pipelines in GPS-denied conditions.

*1) Visual–Inertial Odometry:* We evaluate VINSFusion [23] under low- and high-motion conditions induced by aggressive maneuvers and wind disturbances.

TABLE IV
VISUAL–INERTIAL ODOMETRY PERFORMANCE

| Metric | Low Motion | High Motion |
|---|---|---|
| RMSE (m) | 0.28±0.05 | 0.52±0.12 |
| Drift (m/s) | 0.04±0.01 | 0.09±0.03 |
| Runtime (ms) | 22.1±1.8 | 24.3±2.1 |

*2) LiDAR–Inertial Odometry:* We evaluate FASTLIO2 [36] using the simulated LiDAR and IMU data.

These results indicate that FWAV-Sim provides sensor data with sufficient realism and temporal consistency to support modern odometry pipelines, enabling evaluation of stateestimation performance under aggressive motion and turbulent airflow.

TABLE V
LiDAR–INERTIAL ODOMETRY PERFORMANCE

| Metric | Low Motion | High Motion |
|---|---|---|
| RMSE (m) | 0.15±0.03 | 0.28±0.07 |
| Drift (m/s) | 0.02±0.01 | 0.05±0.02 |
| Runtime (ms) | 35.2±2.8 | 38.1±3.2 |

## V. Conclusion

In this paper, we presented FWAV-Sim, a high-fidelity simulation platform designed to model the coupled interaction between flapping-wing aerial vehicles and turbulent wind environments. By integrating a physics-based aerodynamic solver with spatiotemporally varying wind fields generated via fractal noise, our proposed framework captures key characteristics of flapping-wing flight that are often neglected in existing simulators, including strong aerodynamic nonlinearity, high realistic sensor suit, and stochastic environmental disturbances. Our platform combines quasi-steady wing aerodynamics with a bluff-body drag formulation, enabling realistic force generation on both lifting surfaces and the vehicle body while maintaining computational efficiency suitable for interactive simulation. A central contribution of FWAV-Sim is its dynamic wind manager, which produces spatially correlated, timevarying gusts that can be systematically controlled and reproduced, providing a principled testbed for evaluating robustness under diverse disturbance conditions. Through extensive experiments spanning perception, control, and state estimation, we demonstrated that FWAV-Sim supports modern robotics pipelines and exposes meaningful performance differences across algorithms as environmental complexity increases. The simulator enables benchmarking of perception systems using RGB and LiDAR sensors, evaluation of control strategies under increasing turbulence, and testing of visual–inertial and LiDAR–inertial odometry in GPS-denied scenarios. These results validate FWAV-Sim as a versatile and realistic environment for FWAV research rather than a narrowly scoped aerodynamic model. FWAV-Sim is intended as an enabling infrastructure for the development, analysis, and validation of flapping-wing aerial systems. By narrowing the gap between simplified simulation assumptions and real-world aerodynamic conditions, our platform facilitates more reliable simulationbased development and improves the prospects for simulationto-real transfer. Future work will focus on extending the aerodynamic models to capture additional unsteady effects, incorporating more complex environments, and validating simulation results through comparison with physical FWAV experiments.

## References

[1] Amado Antonini, Winter Guerra, Varun Murali, Thomas Sayre-McCord, and Sertac Karaman. The blackbird dataset: A large-scale dataset for uav perception in aggressive flight, 2018. URL https://arxiv.org/abs/1810.01987.

[2] Muhammad Yousaf Bhatti, Sang-Gil Lee, and Jae-Hung Han. Dynamic stability and flight control of biomimetic flapping-wing micro air vehicle. *Aerospace*, 8(12), 2021. ISSN 2226-4310. doi: 10.3390/aerospace8120362. URL https://www.mdpi.com/2226-4310/8/12/362.

[3] Michael Burri, Janosch Nikolic, Pascal Gohl, Thomas Schneider, Joern Rehder, Sammy Omari, Markus W Achtelik, and Roland Siegwart. The euroc micro aerial vehicle datasets. *The International Journal of Robotics Research*, 35(10):1157–1163, 2016. doi: 10.1177/0278364915620033. URL https://doi.org/10.1177/0278364915620033.

[4] Liang-Chieh Chen, Yukun Zhu, George Papandreou, Florian Schroff, and Hartwig Adam. Encoder-decoder with atrous separable convolution for semantic image segmentation, 2018. URL https://arxiv.org/abs/1802.02611.

[5] Moldoveanu Cristian-Emil, Moraru Florentin, Somoiag Pamfil, Henri Boisson, and Andre Giovannini. Turbulence modelling applied to aircraft wake vortex study. *MTA Review*, XXI, 01 2011.

[6] G.C.H.E. de Croon, K.M.E. de Clercq, R. Ruijsink, B. Remes, and C. de Wagter. Design, aerodynamics, and vision-based control of the delfly. *International Journal of Micro Air Vehicles*, 1(2):71–97, 2009. doi: 10.1260/175682909789498288. URL https://doi.org/10.1260/175682909789498288.

[7] Jeffrey Delmerico, Titus Cieslewski, Henri Rebecq, Matthias Faessler, and Davide Scaramuzza. Are we ready for autonomous drone racing? the UZH-FPV drone racing dataset. In *IEEE Int. Conf. Robot. Autom. (ICRA)*, 2019.

[8] Charles Porter Ellington. The aerodynamics of hovering insect flight. i. the quasi-steady analysis. *Philosophical Transactions of the Royal Society of London. B, Biological Sciences*, 305(1122):1–15, 02 1984. ISSN 0080-4622. doi: 10.1098/rstb.1984.0049. URL https://doi.org/10.1098/rstb.1984.0049.

[9] Fan Fei, Zhan Tu, Yilun Yang, Jian Zhang, and Xinyan Deng. Flappy hummingbird: An open source dynamic simulation of flapping wing robots and animals, 2019. URL https://arxiv.org/abs/1902.09628.

[10] Winter Guerra, Ezra Tal, Varun Murali, Gilhyun Ryou, and Sertac Karaman. Flightgoggles: Photorealistic sensor simulation for perception-driven robotics using photogrammetry and virtual reality. In *2019 IEEE/RSJ International Conference on Intelligent Robots and*

*Systems (IROS)*, pages 6941–6948, 2019. doi: 10.1109/IROS40897.2019.8968116.

[11] L Christoffer Johansson. Aerodynamic efficiency explains flapping strategies used by birds. *Proceedings of the National Academy of Sciences*, 121(46): e2410048121, 2024.

[12] Elia Kaufmann, Antonio Loquercio, Rene´ Ranftl, Matthias Muller, Vladlen Koltun, and Davide Scara-¨ muzza. Deep drone acrobatics, 2020. URL https://arxiv.org/abs/2006.05768.

[13] Matthew Keennon, Karl Klingebiel, and Henry Won. *Development of the Nano Hummingbird: A Tailless Flapping Wing Micro Air Vehicle*. doi: 10.2514/6.2012-588. URL https://arc.aiaa.org/doi/abs/10.2514/6.2012-588.

[14] J. M. Kok and J. S. Chahl. A low-cost simulation platform for flapping wing MAVs. In Akhlesh Lakhtakia, Mato Knez, and Raul J. Mart´ ´ın-Palma, editors, *Bioinspiration, Biomimetics, and Bioreplication 2015*, volume 9429, page 94290L. International Society for Optics and Photonics, SPIE, 2015. doi: 10.1117/12.2084142. URL https://doi.org/10.1117/12.2084142.

[15] Alex H. Lang, Sourabh Vora, Holger Caesar, Lubing Zhou, Jiong Yang, and Oscar Beijbom. Pointpillars: Fast encoders for object detection from point clouds, 2019. URL https://arxiv.org/abs/1812.05784.

[16] Jonggu Lee, Seungwan Ryu, and H Jin Kim. Stable flight of a flapping-wing micro air vehicle under wind disturbance. *IEEE Robotics and Automation Letters*, 5 (4):5685–5692, 2020.

[17] Antonio Loquercio, Elia Kaufmann, Rene Ranftl, Alexey´ Dosovitskiy, Vladlen Koltun, and Davide Scaramuzza. Deep drone racing: From simulation to reality with domain randomization. *IEEE Transactions on Robotics*, 36(1):1–14, 2020. doi: 10.1109/TRO.2019.2942989.

[18] Kevin Y. Ma, Pakpong Chirarattananon, Sawyer B. Fuller, and Robert J. Wood. Controlled flight of a biologically inspired, insect-scale robot. *Science*, 340(6132):603–607, 2013. doi: 10.1126/science. 1231806. URL https://www.science.org/doi/abs/10.1126/science.1231806.

[19] Victor Mwenda Mwongera. A review of flapping wing mav modelling. *Int. J. Aeronaut. Sci. Aerosp. Res*, 2(2): 27–36, 2015.

[20] Quoc-Viet Nguyen and Woei Leong Chan. Development and flight performance of a biologically-inspired tailless flapping-wing micro air vehicle with wing stroke plane modulation. *Bioinspiration & biomimetics*, 14(1): 016015, 2018.

[21] Christopher T. Orlowski and Anouck R. Girard. Dynamics, stability, and control analyses of flapping wing micro-air vehicles. *Progress in Aerospace Sciences*, 51:18–30, 2012. ISSN 0376-0421. doi: https://doi.org/10.1016/j.paerosci.2012.01.001. URL https://www.sciencedirect.com/science/article/pii/S0376042112000103.

[22] Michael O'Connell, Guanya Shi, Xichen Shi, Kamyar Azizzadenesheli, Anima Anandkumar, Yisong Yue, and Soon-Jo Chung. Neural-fly enables rapid learning for agile flight in strong winds. *Science Robotics*, 7(66): eabm6597, 2022.

[23] Tong Qin, Peiliang Li, and Shaojie Shen. Vins-mono: A robust and versatile monocular visual-inertial state estimator. *IEEE transactions on robotics*, 34(4):1004–1020, 2018.

[24] Juan Pablo Rodr´ıguez-Gomez, Raul Tapia, Julio L´ Paneque, Pedro Grau, Augusto Gomez´ Egu´ıluz, Jose Ramiro Mart´ınez-de Dios, and Anibal Ollero. The griffin perception dataset: Bridging the gap between flapping-wing flight and robotic perception. *IEEE Robotics and Automation Letters*, 6(2):1066–1073, 2021.

[25] Olaf Ronneberger, Philipp Fischer, and Thomas Brox. Unet: Convolutional networks for biomedical image segmentation, 2015. URL https://arxiv.org/abs/1505.04597.

[26] Fereshteh Sadeghi and Sergey Levine. Cad2rl: Real single-image flight without a single real image, 2017. URL https://arxiv.org/abs/1611.04201.

[27] Sanjay P. Sane. The aerodynamics of insect flight. *Journal of Experimental Biology*, 206(23):4191–4208, 12 2003. ISSN 0022-0949. doi: 10.1242/jeb.00663. URL https://doi.org/10.1242/jeb.00663.

[28] Shital Shah, Debadeepta Dey, Chris Lovett, and Ashish Kapoor. Airsim: High-fidelity visual and physical simulation for autonomous vehicles, 2017. URL https://arxiv.org/abs/1705.05065.

[29] Shaoshuai Shi, Chaoxu Guo, Li Jiang, Zhe Wang, Jianping Shi, Xiaogang Wang, and Hongsheng Li. Pvrcnn: Point-voxel feature set abstraction for 3d object detection, 2021. URL https://arxiv.org/abs/1912.13192.

[30] Wei Shyy, Hikaru Aono, Chang-kwon Kang, and Hao Liu. *An introduction to flapping wing aerodynamics*, volume 37. Cambridge University Press, 2013.

[31] Haithem Taha, Muhammad R. Hajj, and Ali Nayfeh. Flight dynamics and control of flapping-wing mavs: A review. *Nonlinear Dynamics*, 70, 10 2012. doi: 10.1007/s11071-012-0529-5.

[32] Josh Tobin, Rachel Fong, Alex Ray, Jonas Schneider, Wojciech Zaremba, and Pieter Abbeel. Domain randomization for transferring deep neural networks from simulation to the real world, 2017. URL https://arxiv.org/abs/1703.06907.

[33] Z. Jane Wang. Dissecting insect flight. *Annual Review of Fluid Mechanics*, 37(Volume 37, 2005):183–210, 2005.

ISSN 1545-4479. doi: https://doi.org/10.1146/annurev.fluid.36.050802.121940.
URL https://www.annualreviews.org/content/journals/10.1146/annurev.fluid.36.050802.121940.

[34] Robert J. Wood, Benjamin Finio, Michael Karpelson, Nestor O. Perez-Arancibia, Pratheev Sreetharan, and´ John P. Whitney. Challenges for micro-scale flappingwing micro air vehicles. In Thomas George, M. Saif Islam, and Achyut Dutta, editors, *Micro- and Nanotechnology Sensors, Systems, and Applications IV*, volume 8373, page 83731J. International Society for Optics and Photonics, SPIE, 2012. doi: 10.1117/12.918154. URL https://doi.org/10.1117/12.918154.

[35] Enze Xie, Wenhai Wang, Zhiding Yu, Anima Anandkumar, Jose M. Alvarez, and Ping Luo. Segformer: Simple and efficient design for semantic segmentation with transformers, 2021. URL https://arxiv.org/abs/2105.15203.

[36] Wei Xu, Yixi Cai, Dongjiao He, Jiarong Lin, and Fu Zhang. Fast-lio2: Fast direct lidar-inertial odometry. *IEEE Transactions on Robotics*, 38(4):2053–2073, 2022.

[37] Yan Yan, Yuxing Mao, and Bo Li. Second: Sparsely embedded convolutional detection. *Sensors*, 18(10), 2018. ISSN 1424-8220. doi: 10.3390/s18103337. URL https://www.mdpi.com/1424-8220/18/10/3337.

[38] Tianwei Yin, Xingyi Zhou, and Philipp Krahenb¨ uhl.¨ Center-based 3d object detection and tracking, 2021. URL https://arxiv.org/abs/2006.11275.

[39] Hengshuang Zhao, Jianping Shi, Xiaojuan Qi, Xiaogang Wang, and Jiaya Jia. Pyramid scene parsing network, 2017. URL https://arxiv.org/abs/1612.01105.